# Regularization and Optimization strategies in Deep Convolutional Neural Network

Pushparaja Murugan and Shanmugasundaram Durairaj

*School of Mechanical and Aerospace Engineering,*

*Nanyang Technological University, Singapore 639815*

(`pushpara001@e.ntu.edu.sg, shanmuga005@e.ntu.edu.sg`)

## Abstract

Convolution Neural Networks, known as ConvNets exceptionally perform well in many complex machine learning tasks. The architecture of ConvNets demands the huge and rich amount of data and involves with a vast number of parameters that leads the learning takes to be computationally expensive, slow convergence towards the global minima, trap in local minima with poor predictions. In some cases, architecture overfits the data and make the architecture difficult to generalise for new samples that were not in the training set samples. To address these limitations, many regularization and optimization strategies are developed for the past few years. Also, studies suggested that these techniques significantly increase the performance of the networks as well as reducing the computational cost. In implementing these techniques, one must thoroughly understand the theoretical concept of how this technique works in increasing the expressive power of the networks. This article is intended to provide the theoretical concepts and mathematical formulation of the most commonly used strategies in developing a ConvNet architecture.

## 1 Introduction

Convolutional Neural Networks (CNN) are biologically inspired networks and are developed based on mathematical representation to solve many image visual imagery application, object classification and speech recognition. Hubel and Wiesel studies on cat visual cortex system have inspired the researcher to develop an artificial pattern recognition system. In their study, the proposed that the visual cortex is sensitive to small sub-regions of the visual field, known as a receptive field where simple cells in cortex system respond to edge like patterns and larger receptive fields are invariant to the position of the natural images [1]. Based on this local sensitive and orientation-selective neurons, ConvNets are introduced by Fukushima to solve pattern recognition problem [2] and later developed for identifying the digital characters by LeCun [3]. From this local respective field, neurons are extracted the elementary features and combine them in subsequent layers to get the complex features of the original images. The development of AlexNet in 2012 by Krizhevsky, this structured network is achieved high-level performance and outperformed all other networks in ImageNet- 2012 competition where they trained the 15 million images to classify 22000 classes [4]. There are four main ideas in the development of ConvNet architecture: Local connection, sharing the weights, pooling the data and implementation of multilayers [5]. The typical architecture of ConvNets consists of several layers. There are convolutional layers, pooling layers and fully connected dense layers [6]. The architecture and details are given the Figure 2.1 and Section 2. As a result, many modifications have been developed and implemented in many application.

However, learning of deep network architecture demands a significant amount of data and learning from the input data is a highly computational demanding take. The larger number of connections between the neurons and the parameters are tuned on iterative basis over the cost function by gradient descent optimization or its variants. Also, sometimes developed architecture tends to suffer from overfitting the test data. In this situation, regularization and optimization techniques could be used to overcome the limitation. Computing cost function is a mathematical technique to measure the performance of architecture and to determine the error between actual and predicted values. The gradient descent is optimization technique to determine the optimized values of the parameters to increase the performance and minimize errors in the cost function. Although gradient descent optimization is a natural selection for optimizing the parameters, it has been identified with many limitations on their ability on non-convex complex function and on finding the global minima. Hence, Stochastic Gradient Descent (SGD) and Mini-batch Gradient Descents are developed as the optimization technique to identify the potential solution to find the global minima. Batch gradient decent optimization technique only allows the update of the parameter after the complete computation of gradient value for the entire training samples set. The stochastic gradient descent (SGD) and mini-batch gradient descent optimization allow updating the parameters on each iteration where the updates in SGD carried out on each samples iteration and mini-batch samples iterations. Mini-batch gradient descent is better than SGD on the deep network with highly redundant input data where less computation cost is required for updating the parameters and computation of the gradient can be implemented in GPU as well as in computer cluster. However, the learning rate is fixed in both SGD and mini-batch gradient descent which needs a manual tuning for steady convergence and also to reduce the computational time. Hence, several techniques are developed to accelerate the convergence such as momentum, Nesterov accelerated gradient, Adagrad, AdaDetla, RMS prop, Adam, AdaMax and Nadam.

Regularization technique is used to avoid the overfitting of the network has more parameter than the input data and for the network learned with noisy inputs. Regularization encourages the generalization of the algorithm by avoiding the coefficient to fit so perfectly with the training data samples. To prevent the overfitting, increasing the training sample is an attractive solution. Also, data argumentation, $L_1$ and $L_2$ regularization, Dropout, Drop connect and Early stopping can be used. In ConvNets, increasing the input data, argumentation, early stopping, dropout and its variant are highly implemented for fine-tuning of the network. This article is developed as a continuation of our previous article [7] to address the most commonly used regularization and optimization strategies for developing a typical Convolutional Neural Network framework.

# 2 Architecture

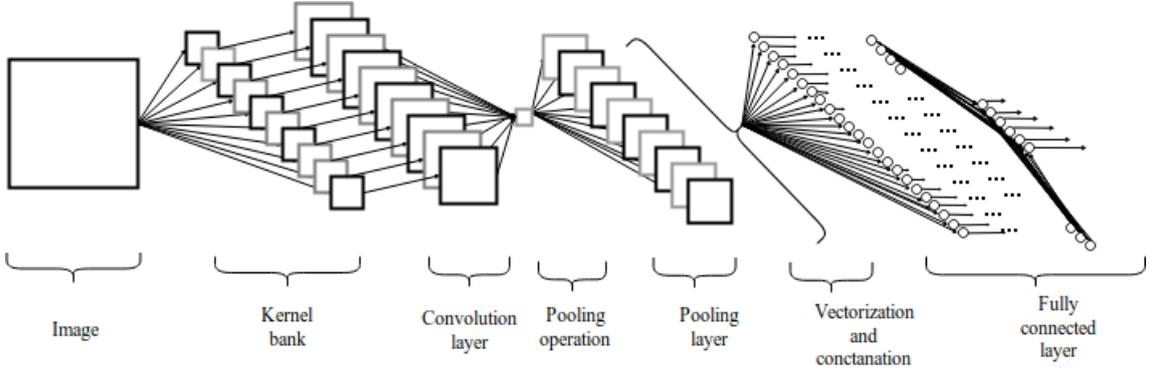

Figure 2.1: Architecture Convolution Neural Network

Set of parallel features maps are developed by sliding different kernels over the input images and stacked together in a layer which is called as Convolutional layer. Using smaller dimension as compares with original image helps the parameters sharing between the feature maps. In the case of overlapping of the kernel with images, zero padding is used to adjust the dimension of the input images. Zero padding also introduce to control the dimensions of the convolutional layer. Activation function decides which neuron should be fired. The weighted sum of input values is passed through the activation layers. The neuron that receives the input with higher values has the higher probability of being fired. Different types of activation function are developed for the past few years that includes linear, Tanh, Sigmoid, ReLu and softmax activation functions. In practice, it is highly recommended that selection of activation function should be based deep learning framework and the field of application. Downsampling of the data is carried out in pooling layers. It reduces the data point and overfitting of the algorithm. Also, pooling layer reduces the noises in the data and smoothing the data. Usually pooling layer is implemented after the convolution process and non-linear transformations. Data points derived from the pooling layers are stretched into single column vectors and fed into the classical deep neural networks. The architecture of a typical ConvNet is given in the Figure 2.1. Cost function, also known as loss function is used to measuring the performance of the architecture by utilizing the actual $y_i$ and predicted values $\hat{y}_i$. Mean Squared Error, Mean Squared Logarithmic Error, $l_1$ $l_2$ norm, Mean Absolute percentage Error and Cross-Entropy are commonly available cost function. In practice, cross entropy is widely used in ConvNet architecture.

The input image patch $z^O$ is convoluted by $D_I^l$ number of maps with size of $C_I^1 \times C_I^l$ and is produced $D_O^l$ number of output maps with the size of $C_O^l \times C_O^l$ where $z^{l-1}$ and $z^l$ are represent the input and output of the layer $l$ and $z^L$ is represent the output of the last layer $L$. The outputs of the layer $l-1$ are fed into the layer $l$ as the inputs. Hence, $j^{th}$ output feature maps of the layer $l$ is denoted by $z_j^l$ and is given as,

$$z_j^l = \sigma(\sum_i z_i^{l-1} * w_{ij}^l + b_j^l 1_{C_o^l}) \tag{Eq. 2.1}$$

where, $0 \leq i < D_I^{(l-1)}, 0 \leq i < D_O^{l-1}$. The convolution process is indicated by the symbol $*$ and the non-linear transformation is represented by $\sigma$. The bias values $b_j^l$ are multiplied with every element of the matrix $1_{C_O^l}$ of size of $C_O^l \times C_O^l$ before added to weighted input values $\sum_i d_i^{l-1} * w_{ij}^l$.

If the down-sampling is implmented in pooling layer by mean pooling, then the $(x, y)$ element ouput of the feature map $j$ of layer $l$ is expressed as,

$$z_j^l(x,y) = \frac{\sum_{m=0}^{s-1} \sum_{n=0}^{s-1} z_j^{l-1}(s \times x + m, s \times y + n)}{s^2} \quad \text{(Eq. 2.2)}$$

where $\leq 0x$, $y < C_I^l$ and $s$ is downsampling factor. In the pooling layer. There are no weights added, and the only parameter is biased. Final layer $L$ is connected to fully connected dense layer where the vectorization and concatenation of the data point of layer $L-1$ are carried out. The output of the last layers is given by, if number of output maps $D_O^L$ for this concatenated layer is $D_O^{L-1} \times (C_O^{L-1})^2$,

$$z_j^L(0,0) = z_i^{L-1}(x,y) \quad \text{(Eq. 2.3)}$$

where,

$$j = i \times (C_O^{L-1})^2 + (y-1) \times C_O^{L-1} + x \quad \text{(Eq. 2.4)}$$

The probability of $m^{th}$ training samples in the class $n \in \{1, 2, \ldots K\}$ for given $K$ number of classses and parameter matrix $\theta = (\theta_1^T, \theta_2^T \ldots \theta_K^T)$ with the size $K \times D_O^L$ is,

$$p(n|z^{L^{(m)}};\theta) = \frac{e^{\theta_n^T d^{L^m}}}{\sum_{c=1}^{K} e^{\theta_n^T d^{L^m}}} \quad \text{(Eq. 2.5)}$$

In practice, softmax classifier is widely used in multi-categorical classification problems. The parameters are identified by using maximum likelihood approach. The cost function Ł for the $m^{th}$ sample belongs to class $n$ is given as,

$$Ł^m = p(n|z^{L^m};\theta) \quad \text{(Eq. 2.6)}$$

If the vector $t$ stores the actual true label is given as $y_i$ and predicted values $z_j^L$ is given as $\hat{y}$ for simpilicty, then the cost function can be written as,

$$Ł(\hat{y}_i, y_i) = \frac{1}{t} \sum_{i=1}^{i=i} f(y_i, \sigma(w_{ij} z_i + b_i)) \quad \text{(Eq. 2.7)}$$

In standard form, the paramters $\theta_i : \{w_i, b_i\}$ that minimize the distance between predicted $\hat{y}_i$ and actual $y_i$ values can be expressed as,

$$Ł(\theta_i) = \frac{1}{i} \sum_{i=1}^{i=i} f(y_i, \hat{y}_i) \quad \text{(Eq. 2.8)}$$

## 3 Gradient descent

Gradient descent, also known as the method of steepest descent, is an iterative optimization algorithm to find the minimum of a complex function by computing successive negative derivatives points of the function through iteration process. The derivative is the rate of change of slope of a function which is typically represented the changes of a function at any given point, provided a continuous and differentiable function. The derivative of a function $f(x)$ with respect to $x$ is $f'(x)$ and is expressed as,

$$f'(x) = \frac{df(x)}{dx} = \lim_{h \to 0} \frac{(x+h) - x}{h} \quad \text{(Eq. 3.1)}$$

From the above Equation 3.1,

- If $f'(x)$ is postive implies that the function is grows up locally,
- If $f'(x)$ is postive implies that the function is grows down locally,
- If $f'(x)$ is postive implies that the function is a stationary point locally,

The gradient descent is given by,

$$x_a = x_{a-1} - \alpha \nabla f(x_{a-1}), \quad a \in \mathbb{R} \tag{Eq. 3.2}$$

where $\alpha$ is the step size.
In case of multivariable function, changes of the function with respect to the points $x, y$ is given by,

$$\frac{\partial}{\partial x} f(x,y) = \lim_{h \to 0} \frac{f(x+h, y) - f(x)}{h} \tag{Eq. 3.3}$$

$$\frac{\partial}{\partial y} f(x,y) = \lim_{h \to 0} \frac{f(x, y+h) - f(y)}{h} \tag{Eq. 3.4}$$

Because of the infinite possibilities of the directions on a multivariable function, the changes of the the function from the points $x, y$ can be represented in vector form as $\vec{v} = <a, b>$, The directional derivatives of the function $f(x, y)$ along the vector $\vec{v}$ at $(x, y)$ can be expressed as,

$$D_{\vec{v}} f(x,y) = \lim_{h \to 0} \frac{f(x+ah, y+bh) - f(x,y)}{h} \tag{Eq. 3.5}$$

The graident desecent is given by,

$$\langle x_a, y_b \rangle = \langle x_{a-1}, y_{a-1} \rangle - \alpha \nabla f(x_{a-1}, y_{a-1}) \tag{Eq. 3.6}$$

Hence, for a given convex and differentiable function $f(x) : \mathbb{R}^n \to \mathbb{R}$, fastest steepest direction $d \in \mathbb{R}$ for a given a point $x \in \mathbb{R}$ towards a neighbours $x = (a_1, a_2..,) \in \mathbb{R}$ exist in the negative derivative of the function $-\nabla f$, if $a < 0 (a \in \mathbb{R})$ such that,

$$f(x + ad) < f(x) \qquad \forall \alpha \in (0, \hat{a}) \tag{Eq. 3.7}$$

The gradient update is given by,

$$\vec{x_a} = \vec{x_{a-1}} - \alpha \nabla f(\vec{x}_{a-1}) \tag{Eq. 3.8}$$

## 3.1 Batch Gradient descent

Batch gradient descent also is known as vanilla gradient descent; computation is carried out to determining the gradient of the function concerning parameters. If predicted value of the out put $\hat{y}_i = f(z^L, \theta_i), \theta \in \mathbb{R}^n$ of the algorithm is obtained by passing the input value $z \in \mathbb{R}^n$ through the intermediate layers and hidden layers. From the Equations 2.6, 2.7 and 2.8, the cost function associated with that layer is given by,

$$\mathrm{L}(\hat{y}_i, y_i) = \frac{1}{t} \sum_{i=1}^{i=i} f(y_i, \sigma(w_i x_i + b_i)) \tag{Eq. 3.9}$$

where, $\sigma$ is the Non-linear activation function. In standard form, the parameters $\theta_i$ that minimize the distance between predicted $\hat{y}_i$ and actual $y_i$ values can be expressed as,

$$\text{L}(\theta_i) = \frac{1}{i} \sum_{i=1}^{i=i} f(y_i, \hat{y}_i) \qquad \text{(Eq. 3.10)}$$

To minimize the cost function, the parameters are initialized and updated in the following manner,

$$\theta_i = \theta_i - \alpha \frac{\partial}{\partial \theta_i} \text{L}(\theta_i : (y_i, \hat{y}_i)) \qquad \text{(Eq. 3.11)}$$

Where the $\alpha$ is the learning rate. The updates of the parameter are performed for all values of $i = (0, 1, 2 \ldots i)$ in the direction of steepest descent of the cost function L [8]. In batch gradient descent, computation of all gradient points for the whole training sample sets has to be carried out to perform a single update which is considered to be inefficient in large datasets. Though the gradient value is zero at the optimum for convex function, batch gradient descent susceptible to fall into local minima for non-convex functions.

## 3.2 Stochastic Gradient descent

Unlike batch gradient descent, stochastic gradient descent (SGD) known as incremental gradient descent, is to update the parameters $\theta$ subsequentially with every iteration. Thus the name comes online learning. SGD iteration computes the noisy gradient for the of input samples and computes the stochastic approximation of gradient descent optimization resulting in larger variance as compares with batch gradient descent. Also, SGD reduces the computational cost as well as avoid the local minima on non-convex function obtained from large datasets.
If the randomly picked sample is $j$ than the SGD optimization is expressed as,

$$\theta_i = \theta_i - \alpha \frac{\partial}{\partial \theta_i} \text{L}(\theta_i : (y_i^{(j)}, \hat{y}_i^{(j)})) \qquad \text{(Eq. 3.12)}$$

The stochastic approximation depends on the randomly picked samples $j = (1, 2, \ldots j)$. In SGD, choosing the learning rate $\alpha$ is a crucial step since the stochastic approximation gradient is non zero at the optimum. The value of the learning rate typically much smaller than the learning rate of batch gradient descent since the tendency of the larger variance in the updates which makes the convergence slow. Larger learning rate encourages the algorithm to stop the iteration before reaching optimum. In practice, this can be addressed by using a small constant learning rate in the initial state to initiate stable convergence and halve learning rate as the convergence slows down. A better method is implementing the annealing learning rate and shuffling the samples before each iteration.

## 3.3 Mini batch Gradient descent

Mini-batch gradient descent involves that splitting the training samples into multiple minibatch contains multiple samples instead of a single sample at every iteration. These mini batches are employed in computing the error and updating the parameter. Sum or average of the gradients for the mini batches reduces the variance as compares with stochastic optimization which leads to more stable convergence. Mini-batch gradient descent is commonly used in deep learning models. If the batch sizes are considered to be $i : i + n$, than the mini batch gradient descent is expressed as,

$$\theta_i = \theta_i - \alpha \frac{\partial}{\partial \theta_i}(\theta : y^{i:i+n}, \hat{y}^{i:i+n}) \qquad \text{(Eq. 3.13)}$$

# 4 Regularizations techniques

Since because of the function of being a potential tool for ensuring the generalization of the algorithm, studies on regularization of the algorithm becomes the main research topic in machine learning [9] [10]. Moreover, the regularization becomes very crucial step in the deep learning model that has more parameters than the training data sets. Regularization is a technique to avoids the overfitting of the algorithm and to avoids the overfitting of coefficients to fit so perfectly as model complexity increases. Overfitting often occurs when the algorithm learns the input data along with noises. Over the past few years, verity of methods are proposed and developed for the machine learning algorithm to regularize such as data argumentation, $L_2$ regularization or weight decay, $L_1$ regularization, dropout, drop connect, stochastic pooling and early stopping [11] [12] [13] [14] [11] [4] [15].

## 4.1 Data agumentation

To increasing the performance of the algorithm as well as satisfying the requirements of a large amount of data for the deep learning model, data augmentation is an important tool to be implemented. Data augmentation is a technique to artificially increase the training set by adding transformations or perturbations of the training data without increasing the computational cost. Data augmentation techniques such as flipping the images horizontally or vertically, crop, color jittering, scaling and rotations are commonly used in the visual imagery and image classification applications. In imagenet classification, Krizhevsky et al. [4] are proposed a methodology of implementing PCA to alter the intensity of the RGB color channels on training AlexNet and also they proposed that the PCA approximately captured the notable properties of the images. Bengio et al., proved that the deep architecture benefits more from the data augmentation technique as compares with the shallow network [16]. Zhang et al., implemented the data argumentation technique along with explicit regularizers such as weight decay and dropouts [15]. Also, it has been used in many image classification problems and found proven as a successful implementation. Chaoyun et al., successfully implemented data augmentation technique to improve the performance of leaf classification and they also claimed that developed ConvNet architecture is outperformed other classification methods [17].

## 4.2 $L_1$ and $L_2$ regularization

The most commonly used regularization methods are $L_1$ and $L_2$ regularization methods. In $L_1$ regularization, regularization term added to the objective function to reduce the sum of absolute value of the parameters wherein $L_2$ regularization, the regularization term is added to reduce the sum of the squares of the parameters. It has been understood from the previous investigations, many parameter vectors in $L_1$ regularization is sparse since the many models cause the parameter to zero. Hence, It has been implemented in feature selection setting where the many features are needed to be avoided or ignored. Most commonly used regularization method in machine learning is imposing a squared $L_2$ norm constraint on the weights. It is also known as weight decay (Tikhonov regularization) since the net effect of reducing the weight by a factor, proportional to the magnitude at every iteration of the gradient descent [18]. Enforcing the sparsity with the weight decay is to artificially introduce the zeros on all lower weights in an absolute manner than referring a threshold. Even in this, the effect of sparsity is negligible. The standard regularized cost function is given by,

$$\theta = \arg\min_{\theta} \frac{1}{N} \sum_{i=1}^{N} (L(\hat{y}_i, y) + \lambda R(w)) \qquad \text{(Eq. 4.1)}$$

where the regularization term $R(w)$ is,

$$R_{L_2}(w) \triangleq ||W||_2^2 \qquad \text{(Eq. 4.2)}$$

Another method is penalize the absolute magintude of the weights, known as $L_1$ regularization.

$$R_{L_1}(w) \triangleq \sum_{k=1}^{Q} ||W||_1 \qquad \text{(Eq. 4.3)}$$

The formulation of $L_1$ regularization is not differentiable at zero, hence weights are increased by a constant factor close to zero. It is common in many neural networks to apply the first order procedures as for the weight decay formulation to solve non-convex $L_1$ regularized problems [19]. The approximate variation of $L_1$ norm is given by,

$$|W|_1 = \sum_{k=1}^{Q} \sqrt{w_k^2 + \epsilon} \qquad \text{(Eq. 4.4)}$$

Another regularization method is considered the mixture of $L_1$ and $L_2$ regularization is known as elastic net penalization [20].

### 4.3 Dropout

Dropout refers to temporarily dropping out the neurons along with their connection. Randomly dropping out the neurons prevent the overfit as well as provide the possible way of combining many different network architecture exponentially and efficiently. Neurons are dropped out with the probability of $1-p$ and reduced the co-adaptability between the neurons. On these stages, the hidden unit usually implemented with a probability of 0.5 drops out neurons. Sample average of all possible $2^n$ drop out neurons are approximately computed by using the full network work with each node's output weighted by a factor of $p$, since using the larger value of $n$ is unfeasible. Drop out significantly reduce the overfitting as well as increasing the learning speed of the algorithm by avoiding the training nodes on training data. The studies and experiments on drop out in fully connected layers by reducing the test result errors from 15.60% to 14.32 % on CIFAR-10 data sets and also implementing dropout in convolution layer reduced error to 12.61%. They also achieved the same trend of improved in the performance on SVH data sets [13]. Hinton et al., implemented the drop out in fully connected layers and they provide that the convolution shared filter architecture reduce the parameters and it improves the resistance to the overfitting in convolutional layers [21]. In AlexNet dropout, regularization technique is used along with data augmentation to reduce the overfitting where they handled 50 million parameters to classify 2200 different categorical images [4].

## 4.4 Drop connect

DropConnect is another regularization strategy to reduce the overfitting of the algorithm which is a generalization of Dropout. In drop connect, sets a randomly selected subset of weights to zero within the architectures instead of a randomly selected subset of activation to zero within each layer. Bounded generalization performance of drop out and drop connect is achieved since each unit receives input from the random subset of previous layers units [22]. Drop connect is similar to the drop out as it involving in introducing sparsity within the model but differs in sparsity of the weights rather than the output vector of the layers.

## 4.5 Early stopping

Early stopping provides the guidance on number passes (iteration) which are needed to be carried out to minimize the cost function. Early stopping is commonly used to prevent the poor generalization of over-expressive models on training. If the number of passes is used is too small, the algorithm tends to underfit (reduce variance but encourage bias) where if the number of passes used is too high (increase variance though reduce bias), algorithm tends to overfit. Early stopping technique is used to address this problem by determining the number of passes and eliminating the manual setting of the values. The fundamental idea behind this early stopping technique is, dividing the available data into three subsets namely, training set, validation set and test set. The training set is used for computing the weights and bias by gradient descent optimization where the validation test is used for monitoring the training process. The computed error tends to decrease as the number of passes increases on both training and validation set. However, if the algorithm begins to overfitting the data, the validation error starts to increase. Early stopping technique provides the possibilities of stopping the passes (iteration) where the deviation in validation error began and returning the weight and bias values [23]. Early stopping is also used in boosting optimization in non-convex loss function [24] and generalization of boosting algorithms [25].

# 5 Optimization techniques

## 5.1 Momentum

Stochastic gradient descent and mini-batch gradient descent are most commonly used in machine learning algorithm to optimize the cost function. However, the learning of the algorithm sometimes is low on the large-scale applications. The momentum strategy is developed to accelerate the learning process, especially in high curvature. The momentum algorithm utilizes the exponentially decaying moving average of the previous gradient values and tends to fall back in that direction [26]. The algorithm introduces a variable $v$ as a velocity vector at which the parameters continue to move in parameter space. The velocity is set to an exponentially decaying average of the negative gradient. For a given cost function to be minimized, the momentum is expressed as,

$$v_{t+1} = \gamma v + \alpha \frac{\partial}{\partial \theta_i}(\theta_i : y_i, \hat{y}_i) \qquad \text{(Eq. 5.1)}$$

$$\theta_{i+1} = \theta_i - v_t \qquad \text{(Eq. 5.2)}$$

where, $\alpha$ is learning rate and $\gamma \in (0, 1]$ is momentum coefficient. $v$ is the velocity vector, and $\theta$ is parameter has the same direction as the velocity vector. Commonly, Stochastic gradient descent will push down the gradient to the steepest side of the long shallow ravine rather than along the ravine towards the optimum. Hence the SGD algorithm becomes very slow to converge. Momentum is one of the strategies to push the gradient along the ravine towards the optimum. In practice, $\gamma$ is initially set to be 0.5 until the stabilization of initial learning and is increased to 0.9. Also, $\alpha$ is set to be very small since the magnitude of the gradient is larger.

## 5.2 Nesterov accelerated gradient (NAG)

Nesterov accelerated gradient is similar to classical momentum algorithm, first-order optimization but differs in gradient evaluations. In NAG, the gradient is evaluated after the implementation of velocity. The updates of this NAG are given by the parameters and as same as the momentum algorithm with better convergence rate. In batch gradient descent, for a smooth convex function, the rate of convergence of the excess from $1/k$ to $1/k^2$ [27]. However, in SGD, Nesterov accelerated gradient does not improve the rate of convergence. The update of NAG is given by,

$$v_{t+1} = \gamma v_t + \alpha \frac{\partial}{\partial \theta_i}(\theta_i : (y_i, \hat{y}_i) - \gamma v_t)) \tag{Eq. 5.3}$$

$$\theta_{i+1} = \theta_i - v_{t+1} \tag{Eq. 5.4}$$

The momentum coefficient is set to be 0.9. In classical momentum algorithm, the current gradient is computed initially then directed towards the updated accumulated gradient. In contrast, the gradient is directed towards the updated accumulated gradient and correction is carried out. This results in preventing the algorithm from going too fast and increase the responsiveness.

## 5.3 Adagrad

Adagrad is known as adaptive gradient allows the learning rate to adapt based on the parameters and eliminate the necessity of tuning the learning rate. The larger update is performed on infrequent parameters, and the smaller update is performed on frequent parameters. Hence, it becomes the natural candidate for sparse data such as image recognition and Natural Language Processing. However the biggest problem of the adagrad is considered to be, in some cases, learning rate becomes too small and network stops the learning process as the learning rate is monotonically decreasing. In classical momentum algorithm and Nesterov accelerated gradient parameters updates are carried out for all the parameters and same learning rate is used on the learning. In adagrad, the different learning rate is used for every parameter at every iteration.

$$G_{i+1} = G_i + \frac{\partial}{\partial \theta_i}(\theta_i : (y_i, \hat{y}_i) \tag{Eq. 5.5}$$

$$\theta_{i+1} = \theta_i - \frac{\alpha}{\sqrt{G^{k-1} + \epsilon}} \frac{\partial}{\partial \theta_i}(\theta_i : (y_i, \hat{y}_i) \tag{Eq. 5.6}$$

## 5.4 AdaDelta

AdaDetla is developed in such a way that it utilizes the recent historical gradient value to scale the learning rate and also similar to classical momentum algorithm, accumulates the historical updates to accelerate the learning. AdaDelta is sufficiently surpassed the weakness of adagrad learning rate converging to zero. Adadetla limits the accumulated previous squared gradient to a fixed window $w$ instead of accumulating the whole historical gradient values. The running average $E[g^2](t)$ at time $t$ depends on the previous average and current gradient values. Hence, the running average is given by,

$$E[g^2]_{i+1} = \gamma E[g^2]_i + (1-\gamma)g_i^2 \quad \text{(Eq. 5.7)}$$

where $\gamma$ is same as momentum term. In practice, it is set to be around 0.9. From the Equation. 3.13, the SDG updates are given as,

$$\theta_i = \theta_i - \alpha \frac{\partial}{\partial \theta_i} \text{Ł}(\theta_i : (y_i^{(j)}, \hat{y}_i^{(j)})) \quad \text{(Eq. 5.8)}$$

From the Equation 5.6, the update of adagrad is given as,

$$\frac{\partial}{\partial \theta_{i+1}} \text{Ł}(\theta_{i+1} : (y_{i+1}^{(j)}, \hat{y}_{i+1}^{(j)})) = -\frac{\alpha}{\sqrt{G_i} + \epsilon} g_i \quad \text{(Eq. 5.9)}$$

Replacing diagonal matrix $G_i$ with past squared gradient $E[g^2]_t$,

$$\frac{\partial}{\partial \theta_{i+1}} \text{Ł}(\theta_{i+1} : (y_{i+1}^{(j)}, \hat{y}_{i+1}^{(j)})) = -\frac{\eta}{\sqrt{E[g^2]_t} + \epsilon} g_t \quad \text{(Eq. 5.10)}$$

where the denominator is root squared eror of the gradient,

$$\frac{\partial}{\partial \theta_{i+1}} \text{Ł}(\theta_{i+1} : (y_{i+1}^{(j)}, \hat{y}_{i+1}^{(j)})) = -\frac{\eta}{RMS[g]_i + \epsilon} g_i \quad \text{(Eq. 5.11)}$$

Replacing the learning rate $\alpha$ in the previous updates rule with $RMS[\frac{\partial}{\partial \theta_i}\text{Ł}(\theta_i : (y_i^{(j)}, \hat{y}_i^{(j)}))$

$$\frac{\partial}{\partial \theta_{i+1}} \text{Ł}(\theta_{i+1} : (y_{i+1}^{(j)}, \hat{y}_{i+1}^{(j)})) = -\frac{RMS[\frac{\partial}{\partial \theta_i}\text{Ł}(\theta_i : (y_i^{(j)}, \hat{y}_i^{(j)}))]_i}{RMS[g]_i} g_i \quad \text{(Eq. 5.12)}$$

$$\theta_{i+2} = \theta_{i+1} + \frac{\partial}{\partial \theta_{i+1}} \text{Ł}(\theta_{i+1} : (y_{i+1}^{(j)}, \hat{y}_{i+1}^{(j)})) \quad \text{(Eq. 5.13)}$$

## 5.5 RMS prop

RMS prop is similar to the first update vector of Adadelta,

$$E[g^2]_{i+1} = 0.9 E[g^2]_i - 0.1 g_i^2 \quad \text{(Eq. 5.14)}$$

The updates of the RMS prop are given as,

$$\theta_{i+1} = \theta_i - \frac{\alpha}{\sqrt{E[g^2]_i} + \epsilon} \frac{\partial}{\partial \theta_i} \text{Ł}(\theta_i : (y_i, \hat{y}_i) \quad \text{(Eq. 5.15)}$$

In RMS prop, learning rate is divided by exponentially decaying average of squared gradients.

## 5.6 Adam

Adam is known as Adaptive Moment Estimation is another technique to compute the adaptive learning rates for the parameters. Similar to adadetla and RMS prop, Adam optimizers stores the exponentially decaying average of the previous squared gradient $v_i$ but also similar to momentum, it keeps the exponentially decaying average of previous historical gradients $m_i$,

$$m_{i+1} = \beta_1 m_i + (1-\beta_1)g_i \qquad \text{(Eq. 5.16)}$$

$$v_{i+1} = \beta_2 v_i + (1-\beta_2)g_i \qquad \text{(Eq. 5.17)}$$

Where $v_i$ and $m_i$ are the estimations of the second and first moment of the gradient respectively. During the initialization stages, first and second momentum are biased to zero. Hence, the bias-corrected first and second moments are given by,

$$\hat{m}_i = \frac{m_i}{1-\beta_1^i} \qquad \text{(Eq. 5.18)}$$

$$\hat{v}_i = \frac{v_i}{1-\beta_2^i} \qquad \text{(Eq. 5.19)}$$

From the above Equations, the adam update is given as,

$$\theta_{i+1} = \theta_i - \frac{\alpha}{\sqrt{\hat{v}_i}+\epsilon}\hat{m}_i \qquad \text{(Eq. 5.20)}$$

In practice, the values of $\beta_1$, $\beta_2$ and $\epsilon$ are set to be 0.9, 0.999 and $10^{-8}$.

## 5.7 Nadam

Nesterov accelerated Adaptive Momentum Estimator (Nadam) is the combination of NAG and Adam optimizers [28]. If the exponentially decaying average of past historical squared gradients is $v_t$ and the exponentially decaying average of past historical gradient is $m_t$, then the classical momentum update rule is given as,

$$g_i = \frac{\partial}{\partial \theta_i}\text{L}(\theta_i : (y_i, \hat{y}_i)) \qquad \text{(Eq. 5.21)}$$

$$m_i = \gamma m_{i-1} \alpha g_i \qquad \text{(Eq. 5.22)}$$

$$\theta_{i+1} = \theta_i - m_i \qquad \text{(Eq. 5.23)}$$

We need to modify the momentum rule to obtain Nadam optimers. Hence expanding the above Equations,

$$\theta_{i+1} = \theta_i - (\gamma m_{i+1} + \alpha g_i) \qquad \text{(Eq. 5.24)}$$

Modification of NAG is given as,

$$g_i = \frac{\partial}{\partial \theta_i}\text{L}(\theta_i : (y_i, \hat{y}_i) - \gamma m_{i-1}) \qquad \text{(Eq. 5.25)}$$

$$m_i = \gamma m_{i-1} + \alpha g_i \qquad \text{(Eq. 5.26)}$$

$$\theta_{i+1} = \theta_i - m_i \qquad \text{(Eq. 5.27)}$$

Modification of NAG can be carried out at one time for updating the gradient $g_t$ and second time for updating the parameters $\theta_{t+1}$, instead of updating the momentum twice. Hence the momentum vector directly update the parameters can be expressed as,

$$m_i = \gamma m_{i-1} \alpha g_i \qquad \text{(Eq. 5.28)}$$

To add the NAG into Adam, previous momentum vectors are needed to be replaced with current momentum vector. Hence, the Adam update rule is given as, Expanding the above Equation with $\hat{m}$ and $m_t$,

$$\theta_{i+1} = \theta_i - \frac{\alpha}{\sqrt{\hat{v}} + \epsilon} \qquad \text{(Eq. 5.29)}$$

$$\theta_{i+1} = \theta_i - (\gamma m_i + \alpha g_i)(\frac{\beta_1 \hat{m}_{i-1}}{1 - \beta_1^i} + \frac{(1-\beta_1)g_i}{1-\beta_1^i}) \qquad \text{(Eq. 5.30)}$$

$$\theta_{i+1} = \theta_i - (\gamma m_i + \alpha g_i)(\beta_1 \hat{m}_{i-1} + \frac{(1-\beta_1)g_i}{1-\beta_1^i}) \qquad \text{(Eq. 5.31)}$$

Utilizing the bias corrected estimation of momentum vector of the previous time step gives update rule for Nadam optimizer and is given as,

$$\theta_{i+1} = \theta_i - (\gamma m + \alpha g_i)(\beta_1 \hat{m}_{i-1} + \frac{(1-\beta_1)g_i}{1-\beta_1^i}) \qquad \text{(Eq. 5.32)}$$

# 6  Conclusion

In this article, the architecture of ConvNet is explained briefly and regularization & optimization strategies are explained elaborately. From the literature, it is clearly understood that implmentation of regularization and optimization strategies is significantly improving the performance and increasing the rate of convergence. Methods for accelerating the optimization by regulating learning rate are discussed. In practice, cross entropy cost function with softmax is commonly used for multi-categorical classifications. Non-linear transformations are usually carried out by employing ReLu function. However, the hyperparameter such as the number & dimension of the convolutional kernel, number & dimensional of pooling layers, number of fully connected dense layers, number of neurons is not covered in this article. These hyperparameters are determined by running the network with the different set of parameters and found out parameters that perform best on the validation test samples.